\documentclass[runningheads]{llncs}
\usepackage{graphicx}
\usepackage{bm}
\usepackage{xcolor}
\usepackage{pgf}
\usepackage{tikz}
\usepackage{pgfplots}

\usepackage{paralist}
\usepackage{comment}

\usepackage{amsmath}
\usepackage{hyperref}
\usepackage{amssymb}
\usepackage{mathtools}
\usepackage{booktabs}
\usepackage{adjustbox}
\usepackage{tabularx}
\usepackage{tikz}
\usetikzlibrary{arrows, positioning}

\usepackage{graphicx}
\usepackage{array,multirow,graphicx}
\usepackage{float}
\begin{document}
\title{A*Net and NBFNet Learn Negative Patterns on Knowledge Graphs}
%
%
\author{Patrick Betz \and Nathanael Stelzner \and
Christian Meilicke \and \\ Heiner Stuckenschmidt \and Christian Bartelt}
\authorrunning{P. Betz et al.}
%
\institute{University of Mannheim, Germany \\
\email{\{first\_name.last\_name\}@uni-mannheim.de}}

\maketitle              

\begin{abstract}
In this technical report, we investigate the predictive performance differences of a rule-based approach and the GNN architectures NBFNet and A*Net with respect to knowledge graph completion. For the two most common benchmarks, we find that a substantial fraction of the performance difference can be explained by one unique negative pattern on each dataset that is hidden from the rule-based approach. Our findings add a unique perspective on the performance difference of different model classes for knowledge graph completion: Models can achieve a predictive performance advantage by penalizing scores of incorrect facts opposed to providing high scores for correct facts. 

\end{abstract}

\section{Introduction}
A knowledge graph (KG) is a structured representation of a certain real world domain in form of (subject, relation, object) triples. The problem of knowledge graph completion (KGC) aims to infer new triples from a given and incomplete KG. Models need to learn patterns from a given KG and use them to make new fact predictions. One model class for performing KGC is given by rule-based approaches~\cite{meilicke2023largeAnyburl,galarraga2013amie}. They learn symbolic patterns on the KG and are fully explainable as fact predictions originate from human-understandable rules. Moreover, they are frequently shown to be highly competitive in regard to predictive  performance~\cite{rossi2021knowledge,meilicke2023largeAnyburl}. Recently, however, a novel graph neural network (GNN) architecture was introduced that reports significant evaluation improvements. The Neural Bellman Ford Network (NBFNet)~\cite{zhu2021neural} and its successor A*Net~\cite{zhu2024net} use a fact scoring mechanism based on calculating path representations of a source and a candidate entity during message passing. Previous work finds that the GNNs express and learn certain positive rules by this procedure~\cite{zhu2021neural,qiuunderstandingQL-GNN}.  

In this technical report, we investigate the GNNs behaviour on the KG benchmarks FB15K-237~\cite{toutanova2015observedFb15k237} and WN18RR~\cite{dettmers2018convolutionalConvE} and compare it with a rule-based approach~\cite{meilicke2023largeAnyburl,ott2023canonical}. Our findings suggest that approximately half of the performance differences of the model classes can be explained by one simple negative pattern on each dataset, which is hidden from the rule-based approach. On WN18RR, the exploitation of an exclusion rule, derived from the particular structure of 1-to-N relations, significantly boosts the performance. For FB15K-237, the particular benchmark construction induced a structural bias disallowing connections between entity pairs at inference time when they are already connected in the  training graph. Despite the pattern being unrealistic and artificial, it can seemingly be exploited by the GNNs when trained under a particular hyperparameter setting. When the setting is turned off, however, we observe a significant performance decrease.

\begin{figure}
\centering
\begin{tikzpicture}[
scale=0.9,
auto, thick, node distance=2cm, 
every node/.style={font=\small},
follows/.style={very thin, ->, >=stealth'},
questioned/.style={very thin, ->, >=stealth', red},
visits/.style={very thin, dashed, ->, >=stealth'}, 
] 

\node (sandy) at (-2.5,0) {sandy};
\node (lisa) at (0,0) {lisa};
\node (bobby) at (2.5,0) {bobby};
\node (anna) at (5,0) {anna};
\node (mike) at (7.5,0) {mike};
\node (zoo) at (2.5,-2) {zoo};
\node[red, inner sep=1pt] (question) at (2.5,1.5) {?};

\foreach \n in {lisa,sandy,bobby, anna, mike, zoo}{
  \draw (\n) circle (6mm); 
}

\draw[follows, shorten >=7pt, shorten <=2pt] (sandy) to node[above,text=blue, pos=0.45] {follows} (lisa);
\draw[follows, shorten >=2pt, shorten <=7pt] (lisa) to node[above,text=blue, pos=0.55] {follows} (bobby);
\draw[follows, shorten >=5pt, shorten <=3pt] (anna) to node[above,text=blue, pos=0.45] {follows} (mike);

\draw[visits, shorten >=8pt, shorten <=8pt] (sandy) to[bend right=20] node[below, sloped, text=blue, pos=0.45] {visits} (zoo);
\draw[visits, shorten >=8pt, shorten <=10pt] (lisa) to[bend right=20] node[below, sloped, text=blue, pos=0.45] {} (zoo);
\draw[visits, shorten >=13pt, shorten <=10pt] (bobby) to[] node[below, sloped, text=blue, pos=0.45] {} (zoo);
\draw[visits, shorten >=8pt, shorten <=12pt] (anna) to[bend left=20] node[below, sloped, text=blue, pos=0.45] {} (zoo);
\draw[visits, shorten >=8pt, shorten <=10pt] (mike) to[bend left=20] node[below, sloped, text=blue, pos=0.45] {} (zoo);

\draw[questioned, shorten >=8pt, shorten <=2pt] (question) to[] node[midway, above, sloped, text=red] {follows} (anna);

\end{tikzpicture}
\caption{A Graphical representation of the zoo dataset. The task is to determine who follows the node $anna$. The correct answer $bobby$ seems to be most intuitive as it visually closes the gap of the graph. However, it is less obvious which specific pattern in the data might support this conclusion and can be learned by a KGC model.}
\label{fig:zoo}
\end{figure}
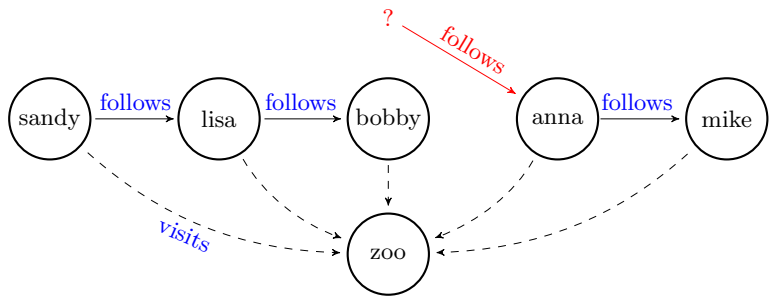

Throughout the following sections, we provide empirical evidence supporting that the identified negative patterns are hidden from the rule-based approach whereas they are exploited by the GNNs. Our experiments are structured into three parts for each dataset. First, we introduce synthetic datasets that allow by their simplicity to identify structural differences between the model classes  (see, e.g., Figure~\ref{fig:zoo}). Subsequently, we perform perturbation-based experiments with respect to the GNNs on the real benchmarks. The procedure is inspired by instance-based explanations and adversarial attacks against KGC models~\cite{bhardwaj-etal-2021-adversarial,pezeshkpour-etal-2019-investigating,LawrenceSN21rollback,betz2022adversarial,wehner2024latentlucidtransformingknowledge}.
We investigate the score changes of GNNs before and after a small perturbation of the training graph. According to our hypothesis, we delete or add facts to activate or deactivate the negative patterns. As the GNNs perform message passing over the training graph during inference time, the analysis can be performed without re-training the models. Finally, we discuss methods of how to easily augment the rule-based approach to also be able to exploit the negative patterns and demonstrate that this leads indeed to performance improvements.

Our findings add a unique perspective to the understanding of the performance differences between the GNNs and other KGC models.  There may be different reasons why a correct prediction is ranked with the highest confidence score. On the one hand, the model may assign it a high score due to potentially learned positive evidence and regularities. On the other hand, the model might simply assign low scores to all other incorrect predictions based on negative patterns learned from the data. In this report, the focus is on the latter.

The remainder of the report is structured as follows. In Section~\ref{sec:background}, we briefly review the required preliminaries. In Section~\ref{sec:synthetic-data}, two synthetic datasets will be discussed and we analyse the model behaviour based on these datasets. Subsequently, the empirical analysis based on WN18RR is presented in Section~\ref{sec:OOT} and for FB15K-237 in Section~\ref{sec:237-OOL}. The last section concludes the work.

\section{Background} \label{sec:background}

\subsection{Knowledge Graph Completion}
We define a KG $\mathcal{G}$ as a collection of $p(s,o)$ facts or triples where $p$ is a relation also termed predicate, $s$ is a head entity, and $o$ is a tail entity. The set of entities is denoted by $\mathcal{E}$ and the set of relations is denoted by $\mathcal{P}$. An existing fact $t \in \mathcal{G}$ is assumed to be a correct statement about the respective domain, such as $profession(obama, politician)$. However, most of the existing real-world KGs are incomplete. A fact which is not existing in the KG is not necessarily false. The field of KGC is concerned with deriving new facts from the KG that are true but were previously unknown. We focus on the standard setting in this report where a model is trained on a training split of a KG and evaluated on a test split of the KG. No additional external data is used, i.e., a model has to learn given patterns from within the training graph and apply them to queries formed from the test graph.

The common evaluation protocols are ranking-based. First, the initial KG is split into training, validation and testing subsets. A model is trained on the training set while using guidance from the validation set. Subsequently, it is evaluated on the test KG. For every fact $p(s,o)$ from the test set, two queries are formed. A head query $p(?, o)$ with correct answer $s$ and a tail query $p(s,?)$ with correct answer $o$. We focus on the tail direction in the following. The explanations are the same for the head direction. A model has to propose confidence scores to candidate answers $c \in \mathcal{E}$ by scoring facts $p(s,c)$. We call $s$ the source entity and $c$ the candidate entity. From the scores of different candidates, a sorted ranking is created and the ranking position of the true answer $o$ is used for evaluation. We use filtered metrics over all experiments. True answers except of the current true answer are filtered out to not penalize models erroneously. Often models calculate directed fact scores, i.e., for some fact $p(e_1, e_2)$ there can be a head score (based on a head query) and a distinct tail score (based on a tail query). We will focus on the common evaluation metrics Mean Reciprocal Rank (MRR) and Hits@X. Further details can be found, for instance, in~\cite{rossi2021knowledge,ruffinelli2020you}.

\subsection{Approaches for KGC} \label{sec:approaches for KGC}

\noindent\textbf{Rule-based KGC.} Rule-based approaches for KGC first mine symbolic Datalog rules on the training KG such as:
\begin{align*}
     &livesIn(X,london) \rightarrow speaks(X,english)  \\
     &bornIn(X,A), \; cityOf(A,Y) \rightarrow citizenOf(X,Y)
\end{align*}
All the variables are universally quantified and $london$ and $english$ are entities of the KG. The comma in the body of the rule denotes the logical \textit{AND}. Rule-based approaches are inherently interpretable as every prediction made can be traced back to a rule which is human readable. However, often a prediction is made simultaneously by multiple rules which requires to define or learn an aggregation function that calculates a final confidence score for the prediction~\cite{galarraga2013amie,betz2024rule,ott2021safran}.

In this report, we focus on approaches that learn aggregation functions on the training graph~\cite{betzSupervised,ott2023canonical}. In particular, we use a simple linear aggregation function as proposed in Ott et al.~\cite{ott2023canonical}. Rules are represented as binary features assigned with learnable weights. Using rules as binary features is also proposed in~\cite{garcia2017kblrn}. Let $t_i$ be some target fact and $\mathbf{x}_i \in \{0,1\}^K$ be a feature vector that represents a learned set of  $K$ rules where $x_{ij}=0$ if rule $j$ predicts $t_i$ and 0 otherwise. The scoring function is based on a simple logistic regression formulation:
\begin{align}
p(t_i|\mathbf{x}_i) = \sigma\bigg(\sum_{j=1}^K w_j x_{ij} + w_0\bigg)  \label{eq:log-reg}
\end{align}
Where the rule weights $\mathbf{w}$ and intercept terms $\mathbf{w}_0$ are learned on the training graph. We follow~\cite{ott2023canonical} and learn a separate model for the tail scores and the head scores and use one intercept term for every relation. For a more detailed overview, we refer to the respective literature. For mining the set of rules, we employ AnyBURL~\cite{meilicke2023largeAnyburl} which uses an anytime algorithm that samples paths on the training KG and generalizes them into certain rule types. \\

\noindent \textbf{Graph Neural Networks.} The introduction of NBFNet and A*Net demonstrated the effectiveness of GNNs for KGC. Compared to the rule-based approach that is used in this work, they achieve an MRR that is roughly 5 percentage points higher on the WN18RR~\cite{dettmers2018convolutionalConvE} and FB15K-237~\cite{toutanova2015observedFb15k237} test sets in the standard setting. Interestingly, the performance is on par when considering the YAGO3-10~\cite{mahdisoltani2013yago3} benchmark~\cite{zhu2024net}. NBFNet and its successor A*Net are message passing GNNs that rely on learning a path representation between a source entity of a query and a candidate entity. In contrast to previous GNN formulations, the message passing is performed in dependence to a current query. By using a boundary condition in the initial round of message passing, all entity representations are initialized with a zero vector except for the source entity of the query. Subsequently, the representations of all connected entities are populated by message passing that originates from the representation of the source entity. The approach learns global relation embeddings but no entity embeddings as their representations are calculated with respect to the current target query. Further details are provided in the original publications. For all our experiments, we use the implementations provided in~\cite{zhu2024net} for NBFNet and A*Net.

\section{Synthetic Datasets} \label{sec:synthetic-data}
In this section, we will present two artificial KGs, the zoo dataset (Section~\ref{sec:zoo dataset}) and the uni dataset (Section~\ref{sec:uni-dataset}). They will help us to identify structural differences between the different model classes. Later in Sections~\ref{sec:OOT} and~\ref{sec:237-OOL}, we will investigate how the model behaviour on the zoo dataset can also be observed on the WN18RR KG benchmark~\cite{dettmers2018convolutionalConvE} and the behaviour on the uni dataset is strongly connected to the particular construction of the FB15K-237 KG~\cite{toutanova2015observedFb15k237}.  

\subsection{The Zoo Dataset} \label{sec:zoo dataset}
Figure~\ref{fig:zoo} shows the zoo KG which only contains two relations and describes students who visit the zoo. They walk in a chain-like structure such that each student has one follower and follows one other student. The dataset that we use in the synthetic experiments described below is exactly as shown in the figure with the exception that we use 100 students nodes and the last student in the chain follows the first student in the chain. Each student visits the zoo (dashed relation) such that message passing over the whole graph is possible. The fact \textbf{\textit{follows(bobby, anna)}} is removed and taken as an evaluation fact, creating a visual gap in the graph. After the training phase on the KG, a model has to propose candidate answers to a head and a tail query formed from the evaluation fact. We will exclusively discuss the head direction within this section for simplicity. Due to the symmetry of the dataset, the same explanations hold for the tail direction for which we also report results for completeness.\\

\noindent \textbf{Experiments on the Zoo Dataset.}
We train different models on the dataset and subsequently compare them by the MRR based on the evaluation fact. The head MRR is based on the single question \textit{"Who follows anna?"} given by the query $follows(?, anna)$; it is calculated from the ranking position of $bobby$ as $\frac{1}{rank(bobby)}$ where the best result of one is achieved when $bobby$ is ranked at the first position. We use a rule-based approach with learned rules from AnyBURL and a trained linear aggregation function~\cite{ott2023canonical} as described in Section~\ref{sec:approaches for KGC}. We compare this to NBFNet and A*Net~\cite{zhu2024net}. We additionally include the knowledge graph embedding (KGE) model ComplEx~\cite{TrouillonWRGB16Complex} based on the libKGE library~\cite{broscheit2020libkge}. During training, no validation set is used and in general the qualitative results do not depend on different hyperparameter configurations. We report averages over 10 independent training runs for all models. \\

\begin{table}[h]
    \setlength{\tabcolsep}{6pt}
    \def\arraystretch{1.2}
    \centering
    \begin{tabular}{ccccc}
        \toprule
        &Approach &MRR Tail & MRR Head & MRR Joint \\
        \midrule
        &Rules (R)& 0.03 $\pm$ 0.02  & 0.02 $\pm$ 0.02 & 0.03 $\pm$ 0.01 \\
        &ComplEx& 0.02 $\pm$ 0.01  & 0.02 $\pm$ 0.02 & 0.02 $\pm$ 0.02 \\
        \hline
        \multirow{2}{*}{\rotatebox[origin=c]{90}{\shortstack{ROH \\ off}}} &NBFNet& 1.00 $\pm$ 0.00  & 1.00 $\pm$ 0.00 & 1.00 $\pm$ 0.00 \\
        &A*Net& 0.93 $\pm$ 0.20  & 1.00 $\pm$ 0.00   & 0.97 $\pm$ 0.10 \\
        \hline
         \multirow{2}{*}{\rotatebox[origin=c]{90}{\shortstack{ROH \\ on}}}&NBFNet& 1.00 $\pm$ 0.00  & 1.00 $\pm$ 0.00 & 1.00 $\pm$ 0.00 \\
        &A*Net& 1.00 $\pm$ 0.00  & 0.95 $\pm$ 0.15   & 0.98 $\pm$ 0.08 \\
        \hline
         &R + $\exists .$-feat.& 1.00 $\pm$ 0.00  & 1.00 $\pm$ 0.00 & 1.00 $\pm$ 0.00 \\
        \bottomrule
    \end{tabular}
    \vspace{0.3cm}
    \caption{Filtered MRR results for the zoo dataset. We report averages ($\pm$ std. dev.) over 10 runs. Random ranking in both directions results in an MRR of approximately 0.02.}
    \label{tab:results-zoo}
\end{table}


\noindent \textbf{Results.} Table~\ref{tab:results-zoo} shows the results. The first two rows describe the rule-based approach and ComplEx, respectively. Rows three to six contain the GNN results. The specifications are identical except for the binary hyperparameter $remove{-}one{-}hop$ (ROH) which does not influence the qualitative results on this dataset and will be discussed in Section~\ref{sec:uni-dataset}. Likewise, the last row will be discussed in Section~\ref{sec:rule-based-exp}. \\

\noindent The rule-based approach and the KGE model are not able to learn anything meaningful for answering the evaluation query. Randomly ranking all nodes would result in an MRR of $\frac{1}{50.5}\approx 0.02$ as there are 100 students plus the $zoo$ node. In fact, the only sensitive behaviour these two model classes show is never scoring the $zoo$ node with the highest likelihood. The GNNs, on the other hand, easily learn to rank $bobby$ at the top position where only A*Net exhibits some noise.\footnote{A fine-grained analysis in which we calculate the the MRR after every training epoch reveals that an MRR (both for head and tail)  of 1 is reached in almost every epoch also for A*Net. In contrast, the rule-based and KGE model never achieve anything higher than reported under various hyperparameter settings.} \\

\noindent \textbf{Why should Bobby be the Follower of Anna?}
Although $bobby$ is an intuitive answer as it visually closes the gap in the training graph, it is not straightforward to formulate which evidence the GNNs might have been used to confidently score $bobby$ higher than all other student nodes. When analysing which nodes typically are followers, we find that everyone who visits the zoo could be a follower. Additionally, everyone who is followed by somebody, is a potential follower by themselves. These observations can help to discriminate $bobby$ from the zoo node (the zoo probably does not follow $anna$) but it does not discriminate $bobby$ from any of the remaining students. $Bobby$ differs from them by the fact that he does not follow anyone (as mentioned, the first student follows the last student in our synthetic dataset). However, this stems from the selection of the evaluation fact in the first place. We cannot draw any conclusion as the data does not contain another student that resembles bobby in this regard.\\

\noindent \textbf{Negative Pattern.} The previous discussion shows that there does not exist positive evidence in the graph (in form of facts) that helps to score $bobby$ with a higher likelihood than the other student nodes. However, the behaviour of the GNNs is quite significant suggesting they learned a non-random regularity. To find an explanation, we have to consider which facts are not present in the graph. For example, $sandy$ does not follow $bobby, \; anna,$ or any of the other nodes but she follows $lisa$. From this we can formulate a negative pattern: \textit{If a node follows someone already, it does not follow someone else.} There is not one case in the data that violates this statement. We will formalize this in Section~\ref{sec:OOT}. In fact, we can easily exploit this pattern and create a scoring mechanism that replicates the behaviour of the GNNs at inference time for the target query. We assign every student node (not the $zoo$ node) a random score from $[0,\epsilon]$, for $\epsilon \in \mathbb{R}_+$, and subtract $\epsilon+1$ whenever a node already follows another node in the training graph. In the end, $bobyy$ will have assigned the highest score and be ranked on top. Remarkably, $bobby$ is not ranked at the first position because he is \textit{ranked up} due to positive evidence, but because the other nodes are \textit{ranked down}.\\

\noindent{\textbf{Discussion.}}
Due to the simplicity of the dataset, there is no other pattern that helps to discriminate all the other students from $bobby$ for the target query except of the fact that they do already follow some other node than $anna$. Therefore, we conjecture that the GNNs exploited this pattern to achieve the performance results. Indeed, when we modify the dataset and let $bobby$ follow some other random student except of $anna$ and subsequently re-train the GNNs, then the MRR collapses.

Intuitively GNNs can exploit the negative pattern as message passing allows a node at inference time to be informed via the follows relation from its successor that it already follows them.  We can view this as a kind of lookup in the training data. Interestingly, although a connection between the source node $anna$ and the candidate node is needed, the pattern is independent of their particular joint path representation. Moreover, the negative pattern will be made explicit in the training data as training involves the construction of negative examples via perturbing the head or tail slot of existing facts. For the rule-based approach, on the other hand, we can be certain that the pattern can not be exploited. The language bias is exactly defined by the allowed rule syntax which does not include negative rules. While KGE models are not the focus of this work, ComplEx shows the same behaviour as the rule-based approach on the dataset. A KGE models does not have access to the training data (in contrast to the GNNs) at inference time, which makes it hard to exploit the negative pattern. In database terms, the $follows$ relationship is, viewed from head-to-tail, a N-to-1 relationship\footnote{Due to the symmetry of the dataset, we even have that N=1 and the relationship is 1-to-1. However that does not affect the results of this section. If we add additional nodes that randomly follow exactly one other node the MRR Head results will be unchanged.} which means that each head entity is associated with exactly one but not more tail entities. In particular, the previous discussion showed how a model might be able to exploit this by the use of a negative pattern formulation. This will be further discussed in Section~\ref{sec:OOT}.

\subsection{The Uni Dataset} \label{sec:uni-dataset}
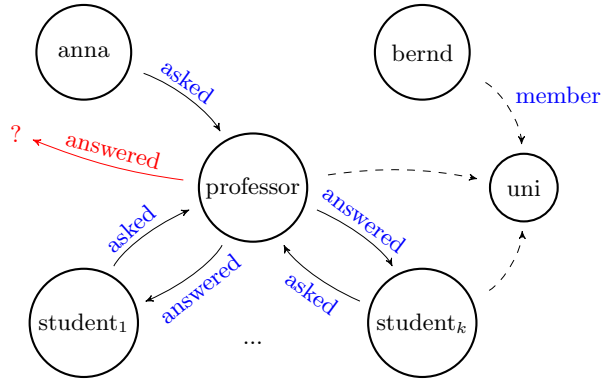
\begin{figure}
\centering
\begin{tikzpicture}[
scale=0.90,
auto, thick, node distance=2cm, 
every node/.style={font=\small}, 
memberOf/.style={very thin, dashed, ->, >=stealth'}, 
asked/.style={very thin, ->, >=stealth'},
answered/.style={very thin, ->, >=stealth'},
questioned/.style={very thin, ->, >=stealth', red}
]

\node (professor) at (0,0) {professor};
\node (anna) at (-2.5,2) {anna};
\node (bernd) at (2.5,2) {bernd};
\node (student_1) at (-2.5,-2) {$\text{student}_1$};
\node (student_i) at (2.5,-2) {$\text{student}_k$};
\node (...) at (0,-2.3) {...};
\node (university) at (4,0) {uni};
\node[red, inner sep=1pt] (question) at (-3.5,0.8) {?};

\foreach \n in {professor,student_1,student_i}{
  \draw (\n) circle (8mm); 
}

\draw (anna) circle (7mm);
\draw (bernd) circle (7mm);
\draw (university) circle (5mm);

\draw[memberOf,shorten >=10pt, shorten <=10pt] (bernd) to[bend left=30] node {\textcolor{blue}{member}} (university);
\draw[memberOf,shorten >=10pt, shorten <=10pt] (student_i) to[bend right=30] (university);
\draw[memberOf,shorten >=8pt, shorten <=7pt] (professor) to[bend left=10] (university);

\draw[asked, shorten >=3pt, shorten <=16pt] (student_1) to[bend left=20] node[pos=0.55,above, sloped, text=blue] {asked} (professor);
\draw[answered,shorten >=3pt, shorten <=16pt] (professor) to[bend left=20] node[pos=0.60,below, sloped, text=blue] {answered} (student_1);

\draw[asked, shorten >=16pt, shorten <=3pt] (student_i) to[bend left=20] node[pos=0.35,below, sloped, text=blue] {asked} (professor);
\draw[answered,shorten >=16pt, shorten <=3pt] (professor) to[bend left=20] node[pos=0.35,above, sloped, text=blue] {answered} (student_i);

\draw[asked,shorten >=16pt, shorten <=10pt] (anna) to[bend left=20] node[pos=0.4,above, sloped, text=blue] {asked} (professor);

\draw[questioned, shorten >=3pt, shorten <=5pt] (professor) to[bend left=7] node[midway, above, sloped, text=red] {answered} (question);
\end{tikzpicture}
\caption{Graphical representation of the uni dataset. Each node is connected via the $member$ relation (dashed arrows) to the $uni$ node. The task is to move the red arrow pointing either to $bernd$ or $anna$. The dataset regularities clearly suggest $anna$; there is no reason why the arrow should be pointing to $bernd$.}
\label{fig:uni-kg}
\end{figure}

Figure~\ref{fig:uni-kg} shows the uni KG where students ask a question to a professor and the professor answers to some of the students. All students (including $anna$ and $bernd$) and the professor are $member$ of the university which is expressed with dashed arrows where some of them are excluded in the figure to prevent occlusion. We duplicate the pattern shown in the figure 3 times, such that there are 3 groups of students and a professor (including 3 versions of $anna$ and $bernd$) but all of them are member of the global $uni$ node.

The fact $answered(professor, anna)$ is removed from the graph and used as an evaluation fact. For simplicity, we will focus on the tail direction in the following. From the evaluation fact, the models have to answer the query $answered(professor, ?)$. The only two relevant candidates are $bernd$ and the correct answer $anna$. The remaining $student_i$ nodes are already known to have received answers from the professor in the training data (they are filtered out by the protocol). When looking at the data, it becomes apparent that $anna$ is the only reasonable choice as there exists a clear positive pattern which can be described by a cyclical rule $asked(X,Y) \rightarrow  answered(X,Y)$ that exhibits support from the $student_i$ nodes. It says that everyone who asked a person also received an answer from that person. While $anna$ asked a question to the professor, nothing is known about $bernd$. Therefore, the data provides clear evidence that $anna$ should be ranked above $bernd$. Conversely, there is no reason why $bernd$ should be ranked above $anna$.

\noindent \textbf{Experiments on the Uni Dataset.} Table~\ref{tab:uni-experiments} shows the results for the evaluation fact. The selected models and details are the same as in the previous section. In the first four rows, all specifications achieve an MRR Tail (and head) of 1.0 after every training run. That is, they all confidently assign the highest score to $anna$ which is expected given the discussion above. \\

\begin{table}[h]
    \setlength{\tabcolsep}{10pt}
    \def\arraystretch{1.2}
    \centering
    \begin{tabular}{ccccc}
        \toprule
        & Approach &MRR Tail & MRR Head & MRR \\
        \midrule
        & Rules& 1.00 $\pm$ 0.00  & 1.00 $\pm$ 0.00 & 1.00 $\pm$ 0.00 \\
        & ComplEx& 1.00 $\pm$ 0.00  & 1.00 $\pm$ 0.00 & 1.00 $\pm$ 0.00 \\
        \hline
         \multirow{2}{*}{\rotatebox[origin=c]{90}{\shortstack{ROH \\ off}}}& NBFNet& 1.00 $\pm$ 0.00  & 1.00 $\pm$ 0.00 & 1.00 $\pm$ 0.00 \\
        & A*Net& 1.00 $\pm$ 0.00  & 1.00 $\pm$ 0.00 & 1.00 $\pm$ 0.00 \\
        \hline
        \multirow{2}{*}{\rotatebox[origin=c]{90}{\shortstack{ROH \\ on}}} & NBFNet&0.22 $\pm$ 0.04&0.33 $\pm$ 0.00& 0.28 $\pm$ 0.02 \\
        & A*Net&0.18 $\pm$ 0.30 & 0.22 $\pm$ 0.13  & 0.20 $\pm$ 0.19 \\
        \bottomrule
    \end{tabular}
    \vspace{0.3cm}
    \caption{Filtered MRR results for the uni dataset. We report averages ($\pm$ std. dev.) over 10 runs.}
    \label{tab:uni-experiments}
\end{table}

\noindent \textbf{Remove-One-Hop (ROH)}. The last two rows of Table~\ref{tab:uni-experiments} show the results for the GNNs when the ROH hyperparameter is turned on. The parameter only affects the training process of the GNNs. During training, for a given training fact $p(s,o)$, all direct connections between entities $s$ and $o$ are removed, i.e., each fact $p'(s,o)$ with $p'\neq p$ is removed from the training graph for the current batch updates.

Recall from Section~\ref{sec:zoo dataset} that for the zoo dataset, the parameter does not have any effect on the experiments. However, for the uni dataset it drastically alters the evaluation results. We pick again the tail direction as an example, but the effects hold for both directions. For instance, the average MRR over 10 runs decreased to 0.18 for A*Net. In fact, in 9 out of 10 runs, $anna$ is assigned a lower score than $bernd$ (and some other student nodes that do not have a direct connection to the professor). For NBFNet, in each of the 10 training runs, $bernd$ is ranked higher than $anna$. \\

\noindent \textbf{Discussion.} It is not straightforward to exactly describe the induced behaviour when the GNNs are trained under the ROH hyperparameter. Nevertheless, the model is never shown a true fact $p(s,o)$ where $s$ and $o$ are directly connected via some other relation. Due to the simplicity of the uni dataset, there is only one edge that distinguishes $bernd$ from $anna$ and it directly connects $anna$ to the professor. Therefore, from the perspective of the query $answered(professor, ?)$, the direct connection between the candidate node  $anna$ and the $source$ node $professor$ must be the reason for the score decrease despite the fact that it actually describes an important positive data regularity. A possibility of replicating the behaviour of the GNNs is given by scoring $anna$ and $bernd$ randomly from $[0,\epsilon]$ and then subtracting $\epsilon+1$ from the score whenever a direct connection between the source node (the professor) and the candidate is given.

\section{Negative Patterns on WN18RR} \label{sec:OOT}
The WN18RR KG was intrudoced by Dettmers et al.~\cite{dettmers2018convolutionalConvE} to have a more challenging version of the WN18 KG~\cite{bordes2013translating}. On the WN18RR test set, the GNNs achieve a joint MRR that is approximately five percentage points higher compared to the rule-based approach (Table~\ref{tab:results-WN18RR}). We have identified a structural difference between the model classes already in Section~\ref{sec:zoo dataset}. The GNNs are able to rank $bobby$ on top for the test query $follows(?, anna)$ by penalizing all remaining student nodes that already follow someone in the training set. We will in the following start with a fine-grained analysis regarding the performance differences of the model classes on the WN18RR dataset. Subsequently, we will introduce the Only-One-Tail (OOT) rule that generalises the regularity that we observed for the zoo dataset. \\

\noindent \textbf{Initial Model Comparison.} We re-trained the GNNs with the implementation from~\cite{zhu2024net} with using the original configuration files and evaluated the models with the evaluation code of the PyClause library~\cite{betz2024pyclause}. There is no noticeable difference to the originally reported results. For the rule-based approach, we use the hyperparameters as provided in~\cite{ott2023canonical}. More details are also described in Section~\ref{sec:approaches for KGC}. 

Results are shown in Table~\ref{tab:results-WN18RR}.  The last rows of the table will be discussed in Section~\ref{sec:rule-based-exp}. The rule-based approach achieves an overall MRR of 0.504 while the GNNs performance is approximately five percentage points higher. Our goal is to find subsets of the data where the performance differences are especially large. Noteworthy, the improvement for the head direction is more than twice as large as for the tail direction.

\begin{table}[h]
    \setlength{\tabcolsep}{3pt}
    \def\arraystretch{1.1}
    \centering
    \begin{tabular}{lccc|c}
    \multicolumn{1}{c}{} & \multicolumn{3}{c}{Overall} & $hypernym$ \\
        \toprule
        Approach &MRR (T) & MRR (H) & MRR \quad  & MRR (H) \\
        \midrule
        Rules (R)& 0.532 & 0.476 & 0.504 & 0.129\\
        NBFNet& 0.561  & 0.541 & 0.551 & 0.274 \\
        A*Net& 0.559& 0.543& 0.551& 0.280\\
        \midrule
         R + $\exists .$-hyper.& 0.530  & 0.520 & 0.525 & 0.238 \\
         R + $\exists .$-all& 0.544  & 0.524 & 0.534 & 0.245 \\
        \bottomrule
    \end{tabular}
    \vspace{0.3cm}
    \caption{Filtered MRR results for WN18RR. MRR (T) and MRR (H) are tail and head MRRs over all relations. The third column is the joint MRR.  }
    \label{tab:results-WN18RR}
\end{table}

\begin{figure}
\centering
\begin{tikzpicture}[scale=0.8]

\begin{axis}[
    axis lines=left,
    ybar=0.0cm,
    bar width=0.5cm,
    enlargelimits=0.15,
    legend style={at={(0.7,1.0)},anchor=north,legend columns=-1},
    ylabel={},
    xtick={2,4,6,8,10,12}, 
    xticklabels={hypernym, d.-rel.-form, mbr-meronym, has-part}, 
    enlarge x limits={abs=1}, 
    xtick=data,
    x tick label style={rotate=35,anchor=east},
    height=6cm,
    width=10cm
]
\addplot[draw=black, fill=blue, area legend] coordinates {(2,1.0) (4,1.0) (6, 1.0) (8,1.0)};
\addlegendentry{NBFNet};
\addplot[draw=black, fill=cyan, area legend] coordinates {(2,1.04) (4,1.0) (6, 0.98) (8,0.92)};
\addlegendentry{A*Net};
\addplot[draw=black, fill=purple, area legend] coordinates {(2,0.48) (4,1.0) (6, 0.93) (8,0.85)};
\addlegendentry{Rules};
\end{axis}
\end{tikzpicture}
\label{fig:rel-results-head-WN18RR}
\vspace{-0.6cm}
\caption{\textbf{Relative relation-wise performance comparison for the head direction}. NBFNet is one. The first four relations with the most facts on the test set are shown (largest first). \textit{Hypernym} consists of 1251 facts while \textit{has-part} only consists of 172 facts.}
\label{fig:relation-wise-WN18RR}
\end{figure}
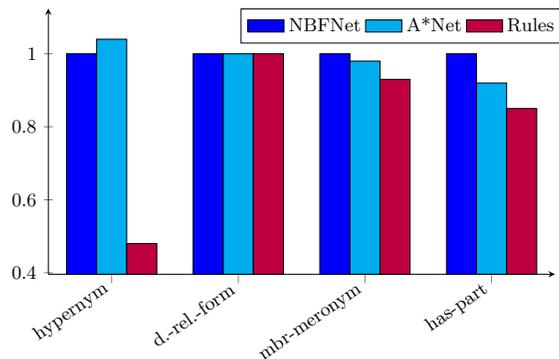

We will therefore have a closer look at the head direction of the performance results. Figure~\ref{fig:relation-wise-WN18RR} shows the relative performance differences (NBFNet is the reference class) of the three approaches for the relations with the most facts in the test set for the head MRRs. WN18RR consists of 11 relations and from the 3134 test facts, 1251 belong to the \textit{hypernym} relation. For the relation with the second most facts (1074) the performance is the same, and the third highest relation only has 253 facts in the test set. The largest effect of the performance differences of the model classes can therefore be pinpointed to the head direction of the \textit{hypernym} relation. To further emphasize this, we calculate a hypothetical MRR of the rule-based approach where we keep all relation and direction results constant but plug in the MRR head results for the \textit{hypernym} relation of NBFNet. This results in a joint MRR of 0.533. \textbf{Therefore, more than 60 percent of the the improvement of the GNNs can be traced back to the improvement with respect to the \textit{hypernym} relation in head direction.} \\

\noindent \textbf{Only-One-Tail Rule}. The \textit{hypernym} relation describes a semantic supertype relationship between a specific and a more generic term, such as \textit{hypernym(red, color)} and \textit{hypernym(blue, color)}. A closer inspection of the WN18RR training set reveals that the relation is of type N-to-1, that is, each head entity can only be associated with exactly one tail entity. While the structural composition of such a relation type can not be explicitly used by the model classes, it can be reformulated to a computational rule, which we express using logic syntax in the following.
\begin{definition}[Only-One-Tail rule]
    Let $x, y, z$ be variables and $p$ be a relation. The Only-One-Tail (OOT) rule w.r.t. $p$ is given by: 
    \begin{align}
        \forall x,y: \quad \exists z \; p(x,z)  \rightarrow  \; \lnot p(x,y) \label{eq:OOT-rule}
    \end{align}
\end{definition}
\noindent The OOT rule is satisfied in the zoo dataset with respect to the $follows$ relationship (you cannot follow $anna$ when you follow someone else already). It is also satisfied for the WN18RR dataset with respect to the $hypernym$ relation as it is a N-to-1 relation. The rule is relevant for the head direction, i.e., for scores based on head queries. It holds within the training graph and it also translates from the training to the test graph. For instance, an entity $c$ cannot be the answer to the test query $hypernym(?, o)$ if it has a $hypernym$ already in the training graph.

Clearly, the GNNs do not execute the negative rule in a crisp fashion as written in~\ref{eq:OOT-rule}. However, we have already seen for the zoo dataset that they are able to exploit such a pattern in contrast to the rule-based approach. Therefore, we hypothesize that one reason for the performance improvements of the GNNs lies in the fact that they are able to penalize candidates that violate the OOT rule.

\subsection{Perturbation Experiments} \label{sec:perturb-exp-WN18RR}
We will investigate in the following if the GNNs exploited the OOT rule on the WN18RR dataset for the $hypernym$ relation. We make use of perturbation experiments in the context of adversarial attacks against KGC models~\cite{betz2022adversarial,pezeshkpour-etal-2019-investigating}. An already trained GNN performs message passing over the training graph for answering test queries at inference times. This allows to query the GNNs based on small perturbations of the training graph. Our experiments are based on the idea that removing or adding a fact that activates or deactivates the OOT rule should be reflected by significant score changes if the rule is indeed exploited by the model. In fact, if the rule is activated, the score of a test fact should decrease while it should increase if the rule is deactivated. \\

\noindent \textbf{Experimental Protocol.}
Generally, all the experiments are independently performed with respect to the 1251 test facts of the \textit{hypernym} relation. First, we define as \textbf{test fact} one of these 1252 facts. We define as \textbf{base fact} the fact on which we perform a score analysis by first scoring it with the original training graph and subsequently scoring it after a perturbation of the training graph. The \textbf{base fact} will be the same as the \textbf{test fact} for the \textit{Add} experiments and it will be derived from the \textbf{test fact} for the \textit{Del} experiments; details are described below. We say that the $\textbf{influential fact}$ is a fact that activates the OOT rule for a base fact. For instance, for the base fact $follows(lisa, anna)$, the influential fact $follows(lisa, bobby)$ exist in the train graph of the zoo dataset while $follows(lisa, mike)$ is an influential fact that does not exist. The $Add$ experiments are based on adding an influential fact when it does not yet exist while the $Del$ experiments are based on removing one that already exists. Further note that the GNNs calculate directed scores. The head score for a fact $r(s,o)$ is based on a query $r(?,o)$ and it is a separate calculation as the tail score for the fact $r(s,o)$ based on a query $r(s,?)$. We will only focus on the head scores in the following given the structure of the OOT rule. \\

\begin{table}[h]
    \setlength{\tabcolsep}{10pt}
    \centering
    \begin{tabular}{ccccc}
        \toprule
         \def\arraystretch{1.2}
        &Model& Avg. Score & $\Delta$ Attack  & $\Delta$ Random  \\
        \midrule
         \multirow{2}{*}{\rotatebox[origin=c]{90}{Add}}&NBFNet &2.39 $\pm$ 1.93 & -3.86 $\pm$ 1.35&-0.64 $\pm$ 0.84 \\
         &A*Net &3.53 $\pm$ 2.59 & -2.89 $\pm$ 1.93 &-0.68 $\pm$ 0.82 \\
        \midrule
         \multirow{2}{*}{\rotatebox[origin=c]{90}{Del}}&NBFNet&-3.94 $\pm$ 0.77 & 2.70 $\pm$ 0.94 & 0.41 $\pm$ 0.54 \\
        &A*Net&-2.79 $\pm$ 1.55 & 1.54 $\pm$ 1.57 & 0.10 $\pm$ 0.55 \\
        \bottomrule
    \end{tabular}
    \vspace{0.3cm}
    \caption{Average head score changes on WN18RR after perturbation experiments with respect to target queries $\_hypernym(?, entity_i)$ based on the test set.}
    \label{tab:score-perturb-WN18RR}
\end{table}

\noindent \textbf{Add.} For each test fact $hypernym(s,o)$, we set the base fact to be the same as the test fact. We calculate the original head score of the this base fact and store it. Subsequently, we select a random entity $o'$ and perturb the training graph by adding the influential fact $hypernym(s,o')$ and calculate the score again. The average difference in scores is termed $\Delta Attack$. We compare this to a random baseline where we sample $o'$ as before but then we also sample a random relation $r' \neq hypernym$ and add the non-influential fact $r'(s, o')$ to the graph. The score difference is termed $\Delta Random$. All perturbations are based on the original training graph such that side effects can be ruled out. \\

\noindent \textbf{Del.} As the OOT rule holds in the dataset, for a test fact $hypernym(s,o)$, the head entity $s$ does not have a $hypernym$ in the training set. In other words, there does not exist a correct fact for which an influential fact regarding the OOT rule with the $hypernym$ relation exists. Therefore, we first have to construct a base fact with an existing influential fact in the training graph. For each test fact $hypernym(s,o)$, we first sample some fact $hypernym(s', e) \in \mathcal{G}^{train}$ with some entities $e,s'\neq s$, which will serve as the influential fact for a constructed base fact. Subsequently, we construct the base fact $hypernym(s', o)$ based on the original object of the test fact. We calculate the score of $hypernym(s', o)$ under the original training graph. Subsequently, we delete $hypernym(s', e)$ from the training graph and calculate the score again. The difference is termed $\Delta Attack$. For the random comparison, we leave the influential fact in the training graph and delete a random triple that contains $s'$ instead. \\

\noindent \textbf{Results.} Table~\ref{tab:score-perturb-WN18RR} shows average score changes and standard deviations over all tested base facts. Adding the influential fact (activating the negative rule) leads to a score decrease of 3.86 for NBFNet and 2.89 for A*Net, respectively. Compared to the averaged initial scores the magnitude of the effects is substantial. For instance, for NBFNet the average score decreases from 2.39 to a negative -1.47. Deleting the influential fact (deactivating the negative rule), on the other hand, leads to a substantial score increase. The effects are significantly stronger compared to the random perturbation although the signs of the effects are the same. In general, it is likely possible that there exist other negative patterns in the dataset which will also affect the random baseline. 

We will further investigate the effects on the performance of the GNNs. For the $Add$ setting, we calculate rankings based on the scores computed after the perturbation. Note that after processing and scoring of a test fact, the graph is reset to the original training graph to exclude side effects. Table~\ref{tab:mrr-perturb-WN18RR} shows the results. The left-hand-side depicts results for the target relation in head direction and the right-hand-side shows overall MRR effects. For all other relations except of $hypernym$, scoring and ranking is unchanged compared to the original model. For NBFNet, the head MRR for $hypernym$ decreases from 0.274 to 0.088 while it decreases from to 0.281 to 0.121 for A*Net. The effects for the overall MRR and Hits values are expectedly smaller as the scoring is unchanged for all other relations. Nevertheless, the MRR decreases from 0.551 to 0.500 for A*Net and to 0.514 for NBFNet. The effects of the random perturbation are also significant, however, the effects are smaller. For instance, the H@1 MRR for the $hypernym$ head direction is decreased from 0.201 to 0.169 for the random perturbation and A*Net while it is decreased to 0.081 under the attack perturbation.

In general, the GNNs comply with the induced behaviour of the OOT rule: The score will decrease if the rule is activated by a perturbation and it will increase if the rule is deactivated for a base fact. Moreover, the observed score changes have significant performance implications.

\begin{table}[h]
   \setlength{\tabcolsep}{5pt}
    \centering
    \def\arraystretch{1.2}
    \begin{tabular}{cc|ccc|ccc}
        \multicolumn{2}{c}{} & \multicolumn{3}{c}{$hypernym$ Head} &\multicolumn{3}{c}{Overall } \\
        \midrule
        & Setting &  MRR &  H@1 & H@10 &   MRR &  H@1 & H@10  \\
        \midrule
        \multirow{3}{*}{\rotatebox[origin=c]{90}{NBFNet}} & Original & 0.274 &  0.197 & 0.428 &   0.551 &  0.496 & 0.661  \\
        & Attack  &  0.088 &  0.066 & 0.131 &  0.500 &  0.458 & 0.586  \\
        & Random  &  0.227 &  0.159 & 0.374 & 0.526   & 0.474   & 0.633   \\
        \midrule
        \multirow{3}{*}{\rotatebox[origin=c]{90}{A*Net}} & Original & 0.281 &  0.207 & 0.420 &   0.551 &  0.498 & 0.655  \\
        & Attack &  0.121 &  0.081 & 0.192 &  0.514 &  0.468 & 0.603  \\
        & Random &  0.242 &  0.169 & 0.371 &  0.535 &  0.482 & 0.638  \\
        \bottomrule
    \end{tabular}
    \vspace{0.3cm}
    \caption{Filtered MRR effects for the head direction of the $hypernym$ relation and overall when adding the influential fact or a random fact individually w.r.t each test fact.}
    \label{tab:mrr-perturb-WN18RR}
\end{table}

\subsection{Rule-based Experiments} \label{sec:rule-based-exp}
For the rule-based experiments, we use a re-implementation of the logistic regression used by Ott et al.~\cite{ott2023canonical} as described in Section~\ref{sec:approaches for KGC}. We discussed already that this approach cannot express the OOT rule for the $hypernym$ relation as the rule cannot be expressed in the language bias. Learning negative rules from the data, on the other hand, is challenging and we are opting here for a simple approach to test our base hypothesis. In fact, we augment equation~\ref{eq:log-reg} by existence features with respect to a current target fact. Let $t =p(s,o)$ be a target fact. We define a binary feature $z(t) \in \{0,1\}$ that evaluates to one, if entity $s$ already exists in the head slot in the KG in some other fact with relation $p$. That is, the feature is one if $\exists \; e \in \mathcal{E}: p(s,e) \in \mathcal{G}$ and it is zero otherwise. First, we create this feature only for the $hypernym$ relation and assign it a learnable weight $\beta$. We then augment the sum in equation~\ref{eq:log-reg} by one term:
\begin{align}
p(t_i|\mathbf{x}_i) = \sigma\bigg(\sum_{j=1}^K w_j x_{ij} + w_0 + \beta z\bigg)  \label{eq:log-reg-augment}
\end{align}
\noindent The additional weight is then learned together with the rule weights using the same training procedure as in~\cite{ott2023canonical}. Note that the new scoring function exclusively affects the scores for the head direction of the $hypernym$ relation. All other scores are unchanged.

We can additionally generalize the concept of existence features and define them for different slots and relation pairs. Although the OOT rule is defined recursively, it is also possible that more general patterns exist that are exploited by the GNNs. For instance, a negative rule could state that an entity cannot exist in the head slot of some relation if it already exists in the tail slot of another relation. We therefore perform an additional experiment where we not only add one term to the sum but we create all possible existence features. We show in the appendix all possible features that can be constructed.

Table~\ref{tab:results-WN18RR} shows the result for WN18RR after training the models. The second last row shows the specification in which we only added one additional feature for the $hypernym$ head direction. This improves the overall MRR already from 0.504 to 0.525. The effect is even more clear when only considering the test facts that are composed of the $hypernym$ relation as only these facts are affected. Here, the result almost doubles from 0.129 to 0.238. Inspecting the learned weight reveals that it is large negative as expected. The last row of the table shows the specification in which we use all possible existence features. We observe an additional improvement with an MRR of 0.534. However, the largest improvement is reached by the single feature for the $hypernym$ head direction. Nevertheless, the results suggest that there exist more negative patterns that can be exploited.

Finally, in Table~\ref{tab:results-zoo}, we perform the same experiment for the zoo dataset in the last row. We add an existence feature that evaluates to one if an entity already follows an entity on the training KG. When re-training the rule-based approach with this additional feature, the behaviour aligns with the GNNs as the learned weight is strongly negative and allows to rank all students down, except of $bobby$.

The experiment results suggest that the initial rule-based approach does not exploit the OOT rule. Adding the existence features, on the other hand, allows it to exploit the negative pattern for the $hypernym$ relation to a large extend.

\section{Negative Pattern on FB15K-237} \label{sec:237-OOL}
We will now turn to the FB15K-237 KG which is frequently used in the KGC community and was introduced in~\cite{toutanova2015observedFb15k237}. The rule-based approach reaches an MRR that is more than 5 percentage points lower compared to the GNNs on the FB15K-237 test set (see Table~\ref{tab:results-237}). We observed in Section~\ref{sec:uni-dataset} for the uni dataset that a direct connection between a source entity ($professor$) and a candidate entity ($anna$) leads to a score decrease for a respective fact when the GNNs are trained under the ROH hyperparameter. In contrast, such a behaviour is not shown by the rule-based approach. In fact, positive rules with only one body atom are important for the overall performance. We will now introduce the Only-One-Link rule:

\begin{definition}[Only-One-Link rule]
    Let $x, y$ be variables and let $\mathcal{P}$ be a set of relations of a KG. Let $\mathcal{P}_c \subseteq \mathcal{P}$ denote a set of (constraint) relations and $p^*\in\mathcal{P}$ be a target relation.  The Only-One-Link (OOL) rule w.r.t. target $p^*$ and constraints $\mathcal{R}_p$ is given by: 
    \begin{align}
        \forall x,y: \bigvee_{p \in \mathcal{P}_b \setminus \{p^*\}}^{}  \; p(x,y)  \rightarrow  \; \lnot p^*(x,y) \label{eq:OOL-rule}
    \end{align}
\end{definition}

\noindent The rule simply says that two entities cannot be connected via a relation $p*$ if they are already connected by a relation from $\mathcal{P}_c$. Unfortunately, the rule does not have relevance for the training set of the FB15K-237 KG alone. Here, it is indeed possible that two entities are directly connected by different relations. However, due to the specific construction of the whole dataset, the rule applies from the training set to the test set for all relations. Two entities $e_1, e_2$ that appear in some fact with any relation, e.g., $p(e_1,e_2) \in \mathcal{G}^{train}$, will never appear together in a fact in the test set by construction~\cite{toutanova2015observedFb15k237}. A model that is able to penalize a candidate score when source and targe entity are already connected in the training graph, will have a significant advantage. Additionally, in our initial ranking analysis, we found that the advantage of GNNs over the rule-based approach was rather evenly distributed over different relations and directions on this dataset.

\subsection{Perturbation Experiments} \label{FB15K-237-perturb-experiements}
In the original configuration files of the GNNs the ROH parameter is turned off for WN18RR and turned on for FB15K-237. We employ perturbation results similar to Section~\ref{sec:perturb-exp-WN18RR} to investigate if the model specifications for FB15K-237 exploit the OOL rule. We re-train the models with the code provided by the original publication~\cite{zhu2024net} and use the evaluation code from the PyClause library~\cite{betz2024pyclause}. We also train and evaluate the models with ROH turned off (Table~\ref{tab:mrr-effects-237}). The perturbation experiments are based on the facts of the test set. The definitions from Section~\ref{sec:perturb-exp-WN18RR} apply, e.g., we term a fact that activates the negative rule an \textbf{influential fact} and the score analysis is based on a \textbf{base fact}. \\

\noindent \textbf{Add.} Every fact from the test set is used as a base fact. For each test fact $p(s,o)$, we calculate the original head and the tail scores. Subsequently, we sample a relation $p'$ and add the influential fact $p'(s,o)$ to the training graph. Then we recalculate the scores for $p(s,o)$. We report the difference of scores before and after the perturbation ($\Delta$ Attack). For the random specification, we sample $p_1', p_2'$, $e_1$ and $e_2$ and add two triples to the graph instead. In particular, we add $p_1'(s, e_1)$ and $p_2'(e_2,o)$ and then recalculate the head and tail score for $p(s,o)$. We do this in favor of the random baseline as the GNNs calculate directed triple scores. \\

\noindent \textbf{Del.} For every test fact $p(s,o)$, we first have to find a base fact for which an influential fact exists. By construction, if $p(s,o)$ is in the test set, then $s,o$ will not appear in another fact in the training set. Therefore, for the tail direction, we sample an existing fact $p'(e,o) \in \mathcal{G}^{train}$ that contains the tail entity of the test fact; it serves as the influential fact. We then create an artificial base fact $p(e,o)$ based on the target relation and the two entities $e$ and $o$. The initial tail score for $p(e,o)$ is calculated, where only facts with a score higher than one are considered to have somewhat realistic artificial facts. Subsequently, we delete the influential fact $p'(e,o)$ from the training set and recalculate the score for $p(e,o)$ leading to the difference in scores. For the random specification, we delete some other triple from the training set instead which contains $o$ but not $e$. Note that the score difference is still based on the artificial base fact $p(e,o)$ before and after the perturbation. For the head direction, we use the same approach but we start with the head entity of the test fact when sampling the influential fact.\\

\noindent \textbf{Results.} Table~\ref{tab:perturb-237} shows average score changes for the tail direction after the perturbation as described above. The results for the head direction are in the appendix and they are qualitatively the same. We observe that activating the OOL rule by adding a direct connection of source and candidate entity leads to a significant decrease in the score. Deactivating the rule by removing an influential fact, on the other hand, leads to an increase. Additionally, the effect of the random specification are marginal although in the $Add$ random setting even two facts are added. The standard deviations that we observe are relatively high. For instance, not in every test case of the $Add$ setting a score decrease is observed. One reason can be that adding the influential fact might induce overlapping effects. Although the OOL is activated it also may activate some other positive pattern that was learned on the training set. However, the experiments clearly suggest that the GNNs have a tendency to penalize a direct connection between the source and the candidate entity.

\begin{table}
\setlength{\tabcolsep}{5pt}
    \centering
    \begin{tabular}{ccccc}
        \toprule
         \def\arraystretch{1.2}
        &Model& Avg. Score & $\Delta$ Attack  & $\Delta$ Random  \\
        \midrule
         \multirow{2}{*}{\rotatebox[origin=c]{90}{Add}}&NBFNet &2.91 $\pm$ 2.88 & -1.78 $\pm$ 2.20 & -0.08 $\pm$ 0.33 \\
         &A*Net &2.89 $\pm$ 3.14 & -2.44 $\pm$ 2.32  &-0.05 $\pm$ 0.68 \\
        \midrule
         \multirow{2}{*}{\rotatebox[origin=c]{90}{Del}}&NBFNet&0.97 $\pm$ 0.92 & 1.49 $\pm$ 1.49 & -0.01 $\pm$ 0.11 \\
        &A*Net&1.07 $\pm$ 1.01 & 1.73 $\pm$ 1.59 & -0.01 $\pm$ 0.06 \\
        \bottomrule
    \end{tabular}
    \vspace{0.3cm}
    \caption{FB15K-237 (Tail direction) perturbation experiments for the default models trained with ROH=on.}
    \label{tab:perturb-237}
\end{table}

\noindent In Table~\ref{tab:mrr-effects-237}, we show the effects on the joint MRR after the $Add$ perturbations. Additionally, we re-train the models with ROH turned off and perform the $Add$ perturbation for these specifications. Note that the $Add$ attack setting essentially leaks the correct answer by the fact that is added to the graph. For instance, for a test fact $p(s,o)$ leading to the tail query $p(s, ?)$ with correct answer $o$, we add $p'(s,o)$ with $p'\neq p$ to the graph. We can view this as informing the model about the correct answer via adding a random connection. Nevertheless, adding the influential fact leads to a significant score decrease as shown above. \textbf{Likewise, we observe that the MRR of NBFNet decreases from 0.416 to 0.287 despite leaking in the correct answer}. Finally, in the right-hand-side of the table, we report the MRR of the models when trained without the ROH parameter. The performance decreases and is only slightly better than the rule-based approach or a good KGE model. Noteworthy, when we repeat the $Add$ experiments for these specifications, the effect turns around. For instance, for NBFNet the MRR increases from 0.371 to 0.454. Intuitively, this behaviour is more natural as informing the model with the correct answer by a random connection should not decrease the ability of the model to find it.

\begin{table}[h]
   \setlength{\tabcolsep}{5pt}
    \centering
    \def\arraystretch{1.2}
    \begin{tabular}{cc|ccc|ccc}
        \multicolumn{2}{c}{} & \multicolumn{3}{c}{ROH=on (default)} &\multicolumn{3}{c}{ROH=off } \\
        \midrule
        Model & Setting &  MRR &  H@1 & H@10 &   MRR &  H@1 & H@10  \\
        \midrule
       \multirow{3}{*}{\rotatebox[origin=c]{90}{NBFNet}} & Original & 0.416  &  0.324 & 0.596 & 0.371 &  0.271 & 0.573  \\
        & Attack  &  0.287 & 0.184 &  0.497 &  0.454 & 0.339 & 0.674  \\
        & Random  &  0.410 &  0.316 & 0.594 & 0.368 & 0.267 & 0.570   \\
        \midrule
        \multirow{3}{*}{\rotatebox[origin=c]{90}{A*Net}} & Original & 0.412 &  0.323 & 0.589 & 0.370&  0.276 & 0.562  \\
         & Attack &  0.256 &  0.178 & 0.410 &  0.406 &  0.291 & 0.635  \\
        & Random &  0.410 &  0.321 & 0.586 &  0.368 &  0.273 & 0.558  \\
        \bottomrule
    \end{tabular}
    \vspace{0.3cm}
    \caption{Filtered MRR and Hits effects on FB15K-237 when adding the influential fact or adding a random fact. Left: The original model specification. Right: The original specification while ROH is turned off. Note that the attack scenario essentially leaks the correct answer by adding a fact $p'(s,o)$ with respect to a test fact $p(s,o)$. For the ROH specification this leads to a decrease in the performance (left) whereas it naturally leads to an improvement when the hyperparameter is turned off (right).}
    \label{tab:mrr-effects-237}
\end{table}

\subsection{Post-hoc Experiments}\label{sec:post-hoc-exp}
We discussed previously that the OOL rule is not reflected within the training set of FB15K-237. Therefore, we are not able to learn weights for existence features on the dataset. While it is possible to add artificial negative facts to the training procedure, we choose the most pragmatic approach that does not involve manipulating the training data. If external knowledge is given about the structure of the OOL rule, we can simply remove all candidate proposals at inference time that already have a connection to the current source entity. By this procedure, we strictly enforce the pattern. For the rule-based approach, we use our re-implementation of the linear model from  Ott et al.~\cite{ott2023canonical} as described in Section~\ref{sec:rule-based-exp} while using their set of rules and hyperparameter configurations. For the post-hoc filter, we choose the following approach exemplified with the tail direction: For a query $p(s,?)$ with correct answer $o$ formed from the test fact $p(s,o)$, we delete every answer candidate $c$ from the ranked list of candidates if $c$ already has a connection to $o$ via some fact in the training set. The approach is the same for the head direction starting with a head query instead. If a model already exploits the OOL rule, the post-hoc filter should not show a significant effect on the evaluation results. On the other hand, models that do not exploit the pattern, should significantly benefit from the filter.

Table~\ref{tab:results-237} contains the results. The first row describes the rule-based approach. The MRR is increased by 2.5 percentage points to 0.385 when applying the post-hoc filter showing a significant improvement. We conclude that half of GNNs performance improvement over a rule-based approach can be explained by exploiting the OOL rule. The next two rows show the GNNs trained under their original hyperparameters, e.g., with ROH turned on. Applying the post-hoc filter has only a marginal effect. On the other hand, in the last two rows, we apply the filter to the GNNs when trained without ROH. In this case, their original MRR is almost completely recovered.

\begin{table}[h]
   \setlength{\tabcolsep}{5pt}
    \centering
    \def\arraystretch{1.2}
    \begin{tabular}{cc|ccc|ccc}
        \multicolumn{2}{c}{} & \multicolumn{3}{c}{Original} &\multicolumn{3}{c}{Post-hoc Filter } \\
        \toprule
        Setting & Approach &  MRR &  H@1 & H@10 &   MRR &  H@1 & H@10  \\
        \toprule
        &Rules & 0.360 &  0.275 & 0.533 & 0.385 &  0.304 & 0.548  \\
        \midrule
         \multirow{2}{*}{\rotatebox[origin=c]{90}{\shortstack{ROH \\ on}}} & NBFNet & 0.416 &  0.324 & 0.596 & 0.417 &  0.325 & 0.597  \\
        & A*Net & 0.412 &  0.324 & 0.589 & 0.415 &  0.325 & 0.592  \\
        \midrule
           \multirow{2}{*}{\rotatebox[origin=c]{90}{\shortstack{ROH \\ off}}} & NBFnet & 0.371 &  0.271 & 0.573 & 0.415 &  0.322 & 0.596  \\
         & A*Net & 0.370 &  0.276 & 0.562 & 0.411  &  0.324 & 0.583  \\
        \bottomrule
    \end{tabular}
    \vspace{0.3cm}
    \caption{Filtered MRR results for FB15K-237 and post-hoc filter experiments for all model classes. The post-hoc filter does almost have no have effect for the original GNN specifications as the OOL rule is already exploited. If ROH is turned off, enforcing the OOL pattern by the filter leads to an increase in performance.}
    \label{tab:results-237}
\end{table}

\noindent In general, the experiments in this section provide further evidence that the rule-based approach is not able to exploit the OOL pattern from the train to the test set of the FB15K-237 KG. On the other hand, the GNNs, when trained with the ROH hyperparameter, seem to have exploited the pattern.

\section{Discussion and Conclusion}
In this report, we compared a rule-based approach~\cite{ott2023canonical} with the GNNs NBFNet~\cite{zhu2021neural} and A*Net~\cite{zhu2024net} with respect to their predictive performance on the benchmarks WN18RR and FB15K-237. We provided empirical evidence that the GNNs are able to exploit certain negative patterns that are hidden from the rule-based approach. Our synthetic experiments also suggest that for these cases, a simple embedding model like ComplEx~\cite{TrouillonWRGB16Complex} aligns with the behaviour of the rule-based approach. 

For WN18RR, our results clearly suggest a higher expressivity of the GNNs compared to the employed rule-based approach. Nevertheless, we showed that simply adding existence features to the rule-based approach leads to a significant performance improvement. The results for FB15K-237 are more ambiguous. It is quite remarkable that the ROH parameter seems to activate the discussed behaviour. It leads to a score penalty for candidates that have already a connection to the source entity of a query. However, the underlying OOL rule is a structural bias introduced by the benchmark construction, opposed to being a natural regularity. It is unclear for a potential user if the models should be trained with the ROH hyperparameter turned on. In general, future work is required to better understand the internal model processes regarding the hyperparameter.

Estimating how much of the overall improvement on the real benchmarks can be explained by the negative patterns is not straightforward. A conservative approach is to base the estimate on the improvement of the rule-based approach under the proposed augmentation strategies. That is, using existence features for WN18RR as in Section~\ref{sec:rule-based-exp} and using the post-hoc filter for FB15K-237 as in Section~\ref{sec:post-hoc-exp}. By doing so, we can be certain that nothing except of the exploitation of the negative patterns is reflected in the improved performance. Therefore, we estimate that approximately half of the improvement of the GNNs over the rule-based approach can be explained by the defined negative patterns.

Finally, it is likely that the patterns that are learned by the GNNs (positive or negative) might overlap. A fact could, for instance, simultaneously activate some positive pattern while also activating some other negative pattern. Indeed, our experiments expectedly suggest tendencies of model behaviour opposed to a hard execution of the negative patterns. In general, we only discussed one particular negative pattern (OOT rule) for one particular relation for WN18RR and one negative pattern regarding all relations for FB15K-237 (OOL rule). It is plausible that the GNNs have learned different negative rules and also more general formulations consisting of distinct relations or even multi-hop patterns. Future work should perform a more fine-grained analysis regarding different pattern types. Moreover, investigating the aggregation of overlapping patterns into a final score is an interesting direction.

\bibliographystyle{splncs04}
\bibliography{references}

\begin{thebibliography}{10}
\providecommand{\url}[1]{\texttt{#1}}
\providecommand{\urlprefix}{URL }
\providecommand{\doi}[1]{https://doi.org/#1}

\bibitem{betz2024pyclause}
Betz, P., Galarraga, L., Ott, S., Meilicke, C., Suchanek, F.M., Stuckenschmidt, H.: Pyclause-simple and efficient rule handling for knowledge graphs. In: IJCAI. Ijcai.org (2024), demo Track

\bibitem{betz2024rule}
Betz, P., L{\"u}dtke, S., Meilicke, C., Stuckenschmidt, H.: Rule confidence aggregation for knowledge graph completion. In: International Joint Conference on Rules and Reasoning. pp. 32--49. Springer (2024)

\bibitem{betz2022adversarial}
Betz, P., Meilicke, C., Stuckenschmidt, H.: Adversarial explanations for knowledge graph embedding models. In: Proceedings of the 31th International Joint Conference on Artificial Intelligence. pp. 2820--2826. Ijcai.org (2022)

\bibitem{betzSupervised}
Betz, P., Meilicke, C., Stuckenschmidt, H.: Supervised knowledge aggregation for knowledge graph completion. In: The Semantic Web. pp. 74--92. Springer International Publishing (2022)

\bibitem{bhardwaj-etal-2021-adversarial}
Bhardwaj, P., Kelleher, J., Costabello, L., O{'}Sullivan, D.: Adversarial attacks on knowledge graph embeddings via instance attribution methods. In: Proceedings of the 2021 Conference on Empirical Methods in Natural Language Processing. pp. 8225--8239. Association for Computational Linguistics (2021)

\bibitem{bordes2013translating}
Bordes, A., Usunier, N., Garcia-Duran, A., Weston, J., Yakhnenko, O.: Translating embeddings for modeling multi-relational data. In: Neural Information Processing Systems (NIPS). pp.~1--9 (2013)

\bibitem{broscheit2020libkge}
Broscheit, S., Ruffinelli, D., Kochsiek, A., Betz, P., Gemulla, R.: Libkge-a knowledge graph embedding library for reproducible research. In: Proceedings of the 2020 Conference on Empirical Methods in Natural Language Processing: System Demonstrations. pp. 165--174 (2020)

\bibitem{dettmers2018convolutionalConvE}
Dettmers, T., Minervini, P., Stenetorp, P., Riedel, S.: Convolutional 2d knowledge graph embeddings. In: Proceedings of the AAAI Conference on Artificial Intelligence. vol.~32, pp. 1811--1818 (2018)

\bibitem{galarraga2013amie}
Gal{\'a}rraga, L.A., Teflioudi, C., Hose, K., Suchanek, F.: Amie: association rule mining under incomplete evidence in ontological knowledge bases. In: Proceedings of the 22nd international conference on World Wide Web. pp. 413--422 (2013)

\bibitem{garcia2017kblrn}
Garc{\'\i}a-Dur{\'a}n, A., Niepert, M.: Kblrn: End-to-end learning of knowledge base representations with latent, relational, and numerical features. UAI  (2018)

\bibitem{LawrenceSN21rollback}
Lawrence, C., Sztyler, T., Niepert, M.: Explaining neural matrix factorization with gradient rollback. In: Thirty-Fifth {AAAI} Conference on Artificial Intelligence, {AAAI} 2021. pp. 4987--4995. {AAAI} Press (2021)

\bibitem{mahdisoltani2013yago3}
Mahdisoltani, F., Biega, J., Suchanek, F.M.: Yago3: A knowledge base from multilingual wikipedias. In: CIDR (2013)

\bibitem{meilicke2023largeAnyburl}
Meilicke, C., Chekol, M.W., Betz, P., Fink, M., Stuckenschmidt, H.: Anytime bottom-up rule learning for large scale knowledge graph completion. The VLDB Journal—The International Journal on Very Large Data Bases  (2023)

\bibitem{ott2023canonical}
Ott, S., Betz, P., Stepanova, D., Gad-Elrab, M.H., Meilicke, C., Stuckenschmidt, H.: Rule-based knowledge graph completion with canonical models. In: Proceedings of the 32nd ACM International Conference on Information and Knowledge Management. pp. 1971--1981 (2023)

\bibitem{ott2021safran}
Ott, S., Meilicke, C., Samwald, M.: {SAFRAN}: An interpretable, rule-based link prediction method outperforming embedding models. In: 3rd Conference on Automated Knowledge Base Construction (2021)

\bibitem{pezeshkpour-etal-2019-investigating}
Pezeshkpour, P., Tian, Y., Singh, S.: Investigating robustness and interpretability of link prediction via adversarial modifications. In: Proceedings of the 2019 Conference of the North {A}merican Chapter of the Association for Computational Linguistics. pp. 3336--3347. Association for Computational Linguistics, Minneapolis, Minnesota (2019)

\bibitem{qiuunderstandingQL-GNN}
Qiu, H., Zhang, Y., Li, Y., et~al.: Understanding expressivity of gnn in rule learning. In: The Twelfth International Conference on Learning Representations (2024)

\bibitem{rossi2021knowledge}
Rossi, A., Barbosa, D., Firmani, D., Matinata, A., Merialdo, P.: Knowledge graph embedding for link prediction: A comparative analysis. ACM Transactions on Knowledge Discovery from Data (TKDD)  \textbf{15}(2),  1--49 (2021)

\bibitem{ruffinelli2020you}
Ruffinelli, D., Broscheit, S., Gemulla, R.: You {CAN} teach an old dog new tricks! on training knowledge graph embeddings. In: International Conference on Learning Representations (2020)

\bibitem{toutanova2015observedFb15k237}
Toutanova, K., Chen, D.: Observed versus latent features for knowledge base and text inference. In: Proceedings of the 3rd workshop on continuous vector space models and their compositionality. pp. 57--66 (2015)

\bibitem{TrouillonWRGB16Complex}
Trouillon, T., Welbl, J., Riedel, S., Gaussier, {\'{E}}., Bouchard, G.: Complex embeddings for simple link prediction. In: Balcan, M., Weinberger, K.Q. (eds.) Proceedings of the 33nd International Conference on Machine Learning. {JMLR} Workshop and Conference Proceedings, vol.~48, pp. 2071--2080. JMLR.org (2016)

\bibitem{wehner2024latentlucidtransformingknowledge}
Wehner, C., Iliopoulou, C., Besold, T.R.: From latent to lucid: Transforming knowledge graph embeddings into interpretable structures (2024)

\bibitem{zhu2024net}
Zhu, Z., Yuan, X., Galkin, M., Xhonneux, L.P., Zhang, M., Gazeau, M., Tang, J.: A* net: A scalable path-based reasoning approach for knowledge graphs. Advances in Neural Information Processing Systems  \textbf{36} (2024)

\bibitem{zhu2021neural}
Zhu, Z., Zhang, Z., Xhonneux, L.P., Tang, J.: Neural bellman-ford networks: A general graph neural network framework for link prediction. Advances in Neural Information Processing Systems  \textbf{34},  29476--29490 (2021)

\end{thebibliography}

\appendix

\section{Appendix}

\subsection{Existence Feature Types for WN18RR}
\begin{table}[h]
    \setlength{\tabcolsep}{10pt}
    \centering
    \begin{tabular}{cccc}
        \toprule
         \def\arraystretch{1.2}
         Target Fact&Type&Condition \\
         \midrule
         $p_j(s,o)$ &head, same slot & $\exists \; e : p_i(s,e) \in \; \mathcal{G}$ \\
         $p_j(s,o)$ &tail, same slot & $\exists \; e : p_i(e,o) \in \; \mathcal{G}$ \\
          $p_j(s,o)$ &head, inverse slot & $\exists \; e : p_i(e,s) \in \; \mathcal{G}$ \\
         $p_j(s,o)$ &tail, inverse slot & $\exists \; e : p_i(o,e) \in \; \mathcal{G}$ \\
         \bottomrule
    \end{tabular}
    \vspace{0.3cm}
    \caption{Possible existence feature types that can be created for one target fact. The overall number of features is 4 times the number of relations. For $p_j=p_i=hypernym$ and type head, same slot, we obtain the representation of the OOT rule.}
\end{table}

\subsection{Perturbation Results for FB15K-237 (Head Direction)}
\begin{table}[H]
    \setlength{\tabcolsep}{10pt}
    \centering
    \begin{tabular}{ccccc}
        \toprule
         \def\arraystretch{1.2}
        &Model& Avg. Score & $\Delta$ Attack  & $\Delta$ Random  \\
        \midrule
         \multirow{2}{*}{\rotatebox[origin=c]{90}{Add}}&NBFNet &1.55 $\pm$ 2.33 & -1.31  $\pm$ 1.88 & -0.05 $\pm$ 0.27 \\
         &A*Net &1.20 $\pm$ 2.44 & -1.90 $\pm$ 1.81 &-0.04 $\pm$ 0.51 \\
        \midrule
         \multirow{2}{*}{\rotatebox[origin=c]{90}{Del}}&NBFNet& 0.82 $\pm$ 0.77 & 1.30 $\pm$ 1.53 & -0.02 $\pm$ 0.12 \\
        &A*Net&0.96 $\pm$ 0.84 & 1.87 $\pm$ 1.49 & -0.01 $\pm$ 0.05 \\
        \bottomrule
    \end{tabular}
    \vspace{0.3cm}
    \caption{FB15K-237 (HEAD direction) perturbation experiments for the default models (ROH=on).}
\end{table}
\end{document}